\documentclass{article}

\usepackage{microtype}
\usepackage{graphicx}
\usepackage{subfigure}
\usepackage{booktabs} % for professional tables

\usepackage{hyperref}

\usepackage[accepted]{icml2019}

\icmltitlerunning{Turn-level Dialog Evaluation with Dialog-level Weak Signals for Bot-Human Hybrid Customer Service Systems}

\def\tightlist{}
\usepackage{graphicx}
\usepackage{amssymb,amsmath}
\usepackage{mathtools}
\usepackage{longtable}
\DeclarePairedDelimiterX{\infdivx}[2]{(}{)}{%
  #1\;\delimsize\|\;#2%
}

\makeatletter
\let\oldlt\longtable
\let\endoldlt\endlongtable
\def\longtable{\@ifnextchar[\longtable@i \longtable@ii}
\def\longtable@i[#1]{
\begin{figure*}[t]
\centering
\onecolumn
\begin{minipage}{\textwidth}
\oldlt[#1]
}
\def\longtable@ii{
\begin{figure*}[t]
\centering
\onecolumn
\begin{minipage}{\textwidth}
\oldlt
}
\def\endlongtable{\endoldlt
\end{minipage}
\twocolumn
\end{figure*}}
\makeatother

\usepackage{enumitem}
\setitemize{noitemsep,topsep=0pt,parsep=0pt,partopsep=0pt}
\begin{document}

\twocolumn[
\icmltitle{Turn-level Dialog Evaluation with Dialog-level Weak Signals \protect\\ for Bot-Human Hybrid Customer Service Systems}

\begin{icmlauthorlist}
\icmlauthor{Ruofeng Wen}{cet}
\end{icmlauthorlist}

\icmlaffiliation{cet}{Self-Service and Automation, CET, Amazon.com}

\icmlcorrespondingauthor{Ruofeng Wen}{ruofeng@amazon.com}
\vskip 0.3in
]

\printAffiliationsAndNotice{} 

\begin{abstract}
We developed a machine learning approach that quantifies multiple
aspects of the \emph{success} or \emph{values} in Customer Service
contacts, at anytime during the interaction. Specifically, the
value/reward function regarding to the turn-level behaviors across human
agents, chatbots and other hybrid dialog systems is characterized by the
incremental information and confidence gain between sentences, based on
the token-level predictions from a multi-task neural network trained
with only weak signals in dialog-level attributes/states. The resulting
model, named Value Profiler, serves as a goal-oriented dialog manager
that enhances conversations by regulating automated decisions with its
reward and state predictions. It supports both real-time monitoring and
scalable offline customer experience evaluation, for both bot- and
human-handled contacts. We show how it improves Amazon customer service
quality in several applications.
\end{abstract}

\hypertarget{background-and-motivation}{%
\section{Background and Motivation}\label{background-and-motivation}}

Customer Service (CS) contacts are commonly studied in the domain of
task-oriented dialogs, where many typical tasks are clear and
well-defined, e.g.~movie and restaurant reservation. These simple tasks
can be abstracted as different API calls, with slots to be filled with
key entities extracted by Natural Language Understanding (NLU) and
Dialog State Tracking (DST) from the conversation. The \emph{success} of
such a dialog is naturally defined as task completion, e.g.~booking
confirmed in the system. However, real life CS contacts for a company
with various lines of products are complex, ambiguous and open-ended. In
fact, many customer issues are manifests of failures in either
functionality or reachability of the current system solutions. As a
result, most CS contacts in real world are handled by human agents, with
assistance from a dialog system, ranging from a simple call handler to
full-fledged chatbots. The task diversity and complexity pose challenges
in these hybrid systems for both dialog evaluation (assess the
\emph{goodness} of bot and agent behaviors) and imitation learning
(train bots or design rules based on \emph{good} agent behaviors).

\begin{figure}
\centering
\includegraphics[width=3.125in,height=2.29167in]{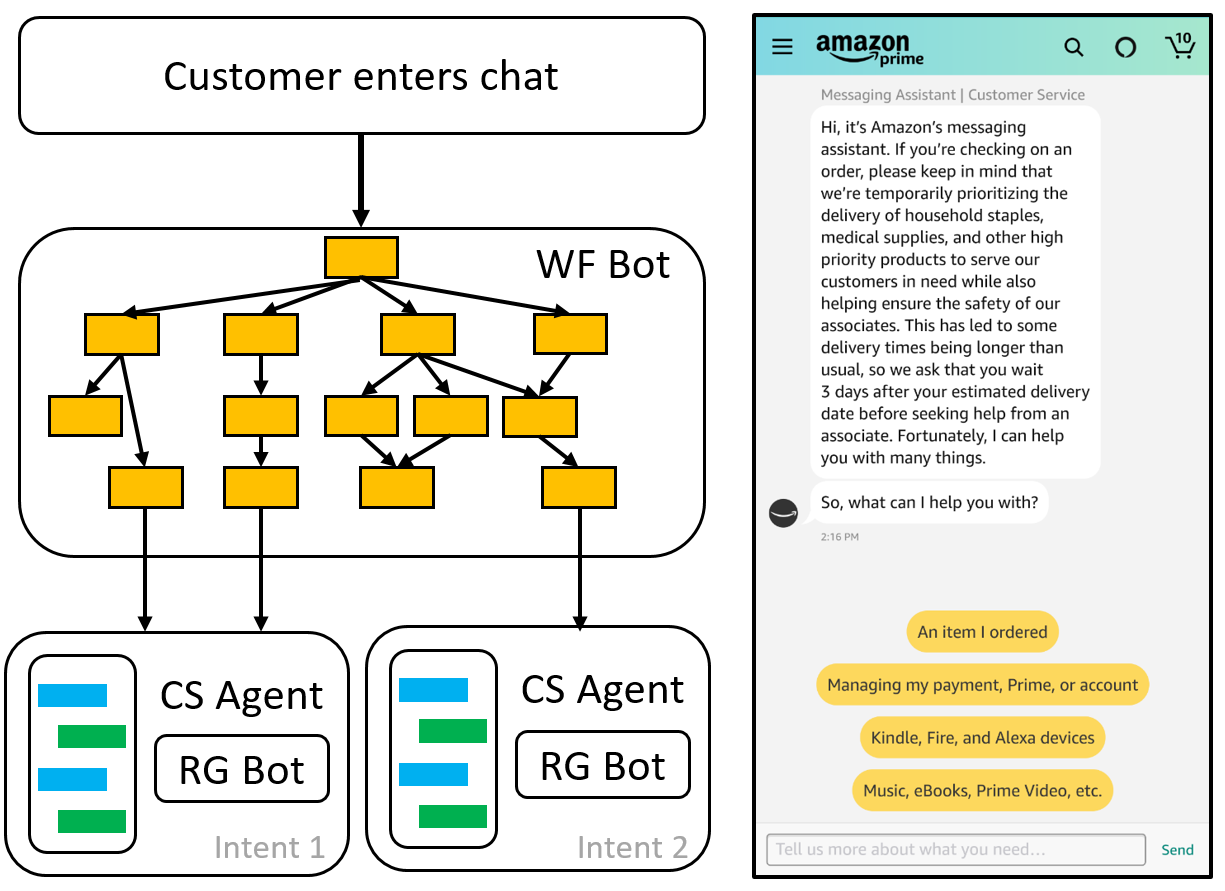}
\caption{Outline of the Amazon CS dialog interface for text chat. Left:
components including WorkFlow (NLU+DST), Response Generator (neural text
generation) and human agents. Right: user interface on Amazon mobile
app.}
\end{figure}

We take the Amazon CS dialog system as an illustrative example for such
hybrid systems, but the discussions are general to any goal-oriented
dialog system with a similar data format. Amazon has a huge amount of CS
contacts every day from world-wide via either phone calls, online chats
or emails, and the level of scale and growth in the number of contacts
demands automated solutions to customer issues in a timely and
satisfactory manner. Amazon's online CS text chat channel is available
to customers on mobile apps and desktops\footnote{You can access it in
  the Amazon mobile app, by \emph{Menu} \(\to\) \emph{Customer Service}
  \(\to\) \emph{Contact Us} \(\to\) \emph{Chat with us}.}. See
\protect\hyperlink{background-and-motivation}{Figure 1} for an overview.
The hybrid dialog system is supported by a combination of rules, machine
learning (ML) powered chatbots and human agents. Customers are first
guided through a `workflow' (WF) bot, consisting of a collection of
designed dialog trees for related customer intents. Different WFs are
triggered by customers clicking preset buttons (e.g.~\emph{questions
about prime membership}) and sometimes entering free texts to describe
their needs. Certain contacts can be auto-resolved by related WFs, while
others are routed to customer service human agents (CSA) with various
skill sets. During the CSA session, a deep learning-based chatbot,
Response Generator (RG;
\href{https://www.aclweb.org/anthology/N19-2007.pdf}{Lu et al., 2018}),
which is trained on historical contact transcripts, recommends a list of
text responses for CSAs to choose from and edit, in order to save typing
and increase efficiency.

Such hybrid systems have drawbacks in each of the components.
\textbf{(1)} There lacks a comprehensive and quantitative understanding
about the success or progress of the complex conversation during the
customer-agent interaction. Although CSAs had proper training, their
responses are not necessarily optimal. Most importantly, there is no
data-driven attribution to actions: \emph{what can we do to actually
improve our service?} \textbf{(2)} Rule-based dialog systems (WF)
require significant human efforts in design and do not scale beyond top
few cases. \textbf{(3)} Dialog generators (RG) learn how to respond by
imitating the most probable past human (CSA) behavior. Intuitively, dull
phrases that do not carry information are prevalent in conversations
under any situation, so they are `safe bets' for the model to predict.
This leads to a preference to generate conversation fillers
(\emph{Thanks}; \emph{A moment please}; \emph{Sure}) instead of
conversation drivers (\emph{I have processed your refund}; \emph{May I
know the reason of the return}). The three problems share a common
solution: the ability to compute and understand at scale the
\emph{values} of any conversation turn and the overall quality.

For evaluating complex dialogs, human annotation is the gold standard
but not scalable. Most automatic metrics for text generation, e.g.~BLEU,
Perplexity, and the recent developed embedding-based ones like BERTScore
(\href{https://arxiv.org/abs/1904.09675}{Zhang et al, 2019}) and
MoverScore (\href{https://arxiv.org/abs/1909.02622}{Zhao et al, 2019}),
are surrogates of text likelihood or similarity, neither measuring the
impact of dialog, nor indicating how to best influence it. In contrast,
business metrics focus on the final impact, not the text itself. For
example, metrics for contact quality may include \emph{no-recontact},
whether the customer calls back in 24 hours, \emph{hows-my-driving}, a
yes-no survey sent after a contact, and \emph{agent-handling-time}.
However, the signals from these metrics are noisy, partial
(e.g.~no-recontact or short handling-time could be a result of customer
frustration) and confounded (e.g.~hows-my-driving survey, if ever
responded, reflects customer's sentiment, not necessarily the quality of
service). These metrics are also sparse signals at dialog-level and not
directly relevant to turn-level actions - the latter is the key to
automatic opportunity seeking and optimization.

For the propensity to generate repeated, dull texts (text degeneration)
of likelihood-based models, it is an active research field with solution
proposals in either model training or decoding
(\href{https://arxiv.org/pdf/1510.03055.pdf}{Li et al., 2015};
\href{https://arxiv.org/pdf/1904.09751.pdf}{Holtzman et al., 2019};
\href{https://arxiv.org/pdf/1908.04319.pdf}{Welleck et al., 2019}).
Though for goal-oriented dialogs, if there exists a well-defined and
strong indicator of success, e.g.~booking confirmed, state-of-the-art
methods can incorporate this reward signal into model training by
Reinforcement Learning techniques, so that the text generator learns to
maximizes success (\href{https://arxiv.org/pdf/1606.02560.pdf}{Zhao and
Eskenazi, 2016}; \href{https://arxiv.org/abs/1609.00777}{Dhingra et al.,
2016}; \href{https://arxiv.org/pdf/1702.03334.pdf}{Kandasamy et al.,
2017}; \href{https://arxiv.org/pdf/1608.05081.pdf}{Lipton et al., 2017};
\href{https://arxiv.org/abs/1712.02838}{Zhou et al., 2017}). However, as
discussed, it is difficult to find such signals for complex dialogs like
CS contacts. Therefore it is necessary to first design a
business-aligned Value/Reward/Critic based on some weak but scalable
signals, in order to train a Policy/Agent/Actor to make decisions that
add most value.

\textbf{Our contribution} This work aims to attack the above challenges
in separate components of a bot-human hybrid CS dialog system, with a
single end-to-end solution. \textbf{(1)} In term of application, we
present Value Profiler (VP), a multi-task neural network that reads
customer contact conversation, assesses the situation and acts in real
time to maximize the newly defined value/success, serving as a
data-driven dialog manager to regulate natural language
understanding/generation and dialog state tracking. \textbf{(2)} As for
methodology, we propose a new formulation to evaluate goal-oriented
dialog agent behaviors at scale and to design turn-level reward
functions for bots, with the presence of only dialog-level attributes as
weak signals.

In \protect\hyperlink{method}{Section 2} we detail the Value Profiler
model, and test its performance with experiments for different
application areas in the Amazon customer service domain in
\protect\hyperlink{applications}{Section 3}. Additional related research
work is reviewed in \protect\hyperlink{related-work}{Section 4}.
\protect\hyperlink{conclusion-and-future-work}{Section 5} concludes the
paper with future plans.

\hypertarget{method}{%
\section{Method}\label{method}}

The methodology is introduced in the following steps: (1) problem
formulation - how to quantify the qualitative criteria of a successful
CS contact, using the available datasets and a predictive model; (2)
predictive model design details; (3) how to define the \emph{values}
based on the predictions and apply in practice.

\hypertarget{problem-formulation-and-data}{%
\subsection{Problem Formulation and
Data}\label{problem-formulation-and-data}}

Without a single gold standard, we argue that a successful customer
service contact is supported by the following aspects:

\begin{itemize}
\tightlist
\item
  the agent understands customer's issue well
\item
  the agent knows which action to take
\item
  the contact outcome satisfies the customer
\item
  the above process induces low costs
\end{itemize}

These criteria can be translated into some equivalent ones in statistics
and ML terms, changing the subject from a human to a bot or model:

\begin{itemize}
\tightlist
\item
  a model predicts issues with high confidence
\item
  a model predicts actions with high confidence
\item
  a model predicts high \(P(\text{no-recontact})\)\footnote{Throughout
    this text, \emph{no-recontact} is used to measure customer
    satisfaction, for demonstration. Note \emph{hows-my-driving} survey
    result is in exactly the same binary yes/no format so all discussion
    applies to it.}
\item
  a model predicts low costs with confidence
\end{itemize}

This formulation translates the assessment of success criteria into a
multi-task supervised learning problem (classification for
issue/action/no-recontact, regression for costs). The model is required
to be calibrated, i.e.~its outputs are proper probability distributions
that faithfully reflect the underlying uncertainties within the data
generation process, so that the model confidence measure is reliable. We
defer the essential discussion of how to compute the level of model
\emph{confidence} and the \emph{value} progress within the dialog to
\protect\hyperlink{value-function}{Section 2.3}, and for now focus on
this predictive model itself.

We briefly describe the high-level data structure need for the this
formulation. Dialog texts need to contain all utterances from bot,
customer and agent, with sensitive information removed. The issue could
simply be a categorical variable, or has multiple layers of hierarchical
tree structure, e.g.~3 layers from the most abstract level
(e.g.~\emph{digital and device}), to the finest grain with more concrete
cases (e.g.~\emph{item received not expected}). For the dataset used in
this paper, the action labels are noisy and not turn-aligned so they are
used also as a dialog-level outcome. Actions can be merged into some
main categories (e.g.~\emph{refund created}) to avoid sparsity.
Recontact within certain time should be well-defined. In this text we
focus on the values in issue/action/recontact, and leave the cost aspect
to future work.

In general, the required datasets are simply conversation texts and
dialog-level outcomes, some of them we have a preference (no recontact,
low costs), some of them we want to be as sure as possible (issue,
action). A different goal-oriented dialog application could plug in its
own available labels, e.g.~dialog states slot values and the downstream
impacts, and all discussions apply.

\begin{figure}
\centering
\includegraphics[width=3.125in,height=1.66667in]{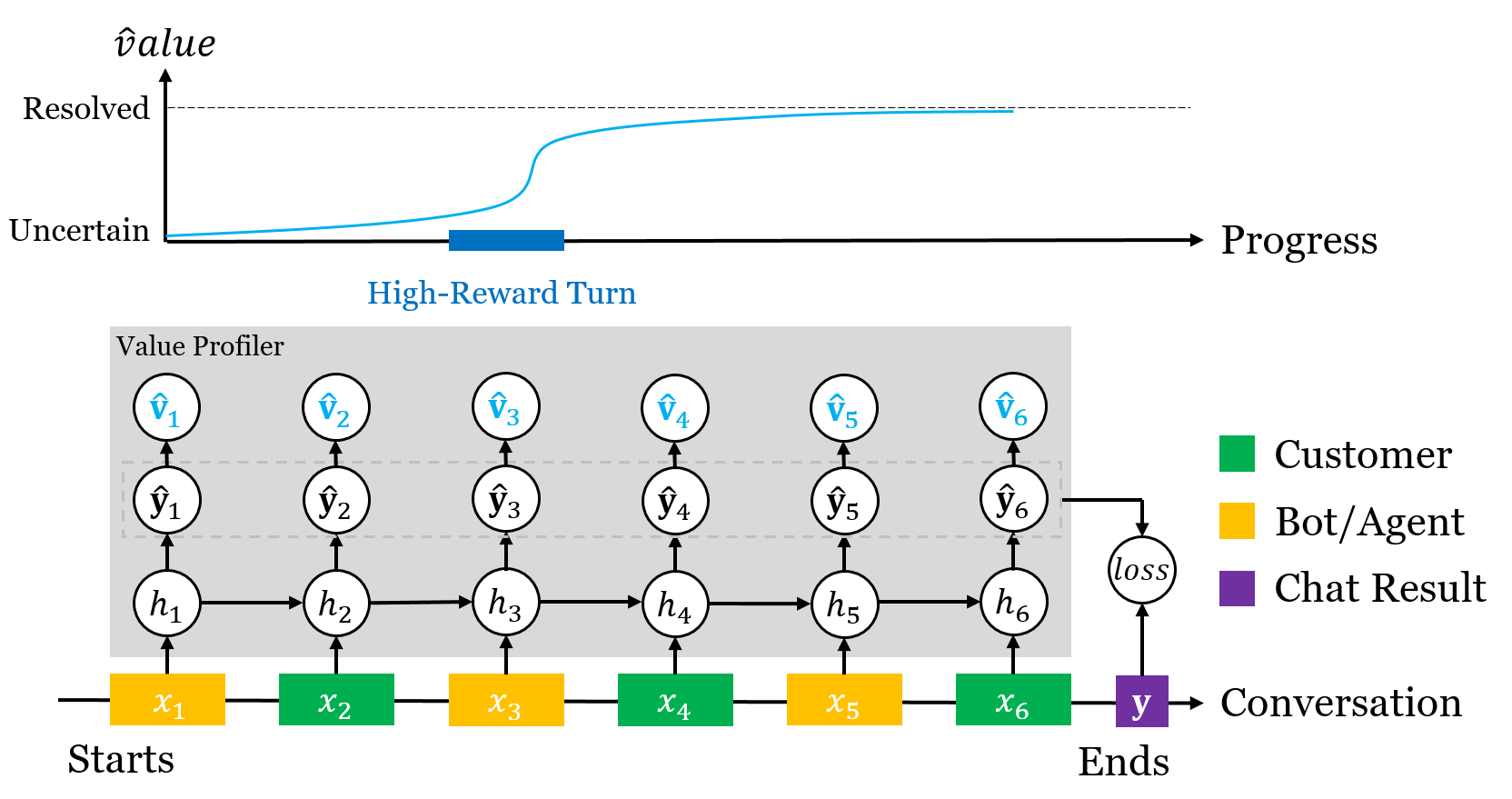}
\caption{Value Profiler Illustration. Bottom: VP encodes conversation
texts into hidden representations, generates predictions for the dialog
outcomes at each turn, and compute \emph{values} (see definition in
\protect\hyperlink{value-function}{Section 2.3}) based on the
predictions. Top: an idealized value progress curve, for illustration.}
\end{figure}

\hypertarget{value-profiler-predictive-model}{%
\subsection{Value Profiler Predictive
Model}\label{value-profiler-predictive-model}}

\begin{longtable}[]{@{}lrr@{}}
\caption{Issue prediction accuracy improved by VP, with 95\% confidence
interval, for top intents. All \% are relative to the WF baseline
accuracy (not shown).}\tabularnewline
\toprule
last WF intent & Acc. improved \% & 95\% CI\tabularnewline
\midrule
\endfirsthead
\toprule
last WF intent & Acc. improved \% & 95\% CI\tabularnewline
\midrule
\endhead
account & 78.13\% & {[}73.26,82.69{]}\tabularnewline
delivered-not-received & -0.03\% & {[}-0.06,0.01{]}\tabularnewline
item-delivered & 1.72\% & {[}1.34,2.07{]}\tabularnewline
item-in-transit & 2.51\% & {[}2.17,2.87{]}\tabularnewline
item-not-received & -0.02\% & {[}-0.05,0.01{]}\tabularnewline
live-help-request & 3.50\% & {[}3.24,3.78{]}\tabularnewline
order-related & 6.18\% & {[}5.55,6.76{]}\tabularnewline
prime & 59.00\% & {[}55.66,62.37{]}\tabularnewline
return-refund-status & -1.97\% & {[}-2.44,-1.5{]}\tabularnewline
returns-refunds & -0.65\% & {[}-1.59,0.39{]}\tabularnewline
start-return & 0.02\% & {[}-0.02,0.07{]}\tabularnewline
unknown-charge & 8.77\% & {[}8.15,9.41{]}\tabularnewline
wms & 0.72\% & {[}0.44,1.01{]}\tabularnewline
all-other-intents & 3.01\% & {[}2.69,3.33{]}\tabularnewline
TOTAL & 3.76\% & {[}3.64,3.88{]}\tabularnewline
\bottomrule
\end{longtable}

Essentially, Value Profiler (VP) is a multi-task (predict issues,
actions, recontact, etc.), token-level (predict at each token)
sequential supervised learning model. See
\protect\hyperlink{value-profiler-predictive-model}{Figure 2} for
illustration. The input to VP is the conversation history in the form of
a sequence of tokens: \(x_{:t,i} = (x_{1,i},\dots,x_{t,i})\) where \(i\)
is dialog index and \(t\) is token index. A causal-masked encoder
\(h(\cdot)\) then process the texts in a left-to-right manner, and
outputs predictions at each token
\(\hat{\mathbf{y}}_{t,i} = h(x_{:t,i})\). The causal masking ensures the
output at certain token only depends on the history, and any encoders
with causal property can apply. In this paper, the pretrained
transformer GPT2
(\href{https://cdn.openai.com/better-language-models/language_models_are_unsupervised_multitask_learners.pdf}{Radford
et al., 2019}) is used and benchmarked with two common choices: LSTMs
and dilated causal 1D CNNs
(\href{https://arxiv.org/pdf/1609.03499.pdf}{Oord et al., 2016}).

For the three tasks, VP takes issue (or each leaf node of the issue
tree) as a multi-class classification problems in the common
softmax-cross-entropy setup. The action task is treated as a multi-label
classification problem, consists of a set of binary classifications.
This setup is a result of data structure: most dialogs have only a
single intent, while multiple actions could be performed. Finally
no-recontact is simply another binary classification for customer
satisfaction.

The prediction tasks are at token or sentence-level to enable dialog
evaluation at any point, for any actions presented in the texts. Since
all target signals are dialog-level, they are replicated along the
sequence: \(\mathbf{y}_{t,i} = \mathbf{y}_i\). This effectively trains
all possible prediction scenarios within a conversation at the same time
in a single sample (if length permits):
\(\hat{\mathbf{y}}_{:t,i} = h(x_{:t,i})\) where \(h\) is causal so
\(\hat{\mathbf{y}}_{t-k,i}\) is generated only using the \(x_{:t-k,i}\)
part of the input. The loss for the \(i\)th sample/dialog is defined as
the sum of token-level losses:
\(\sum_t{L(\hat{\mathbf{y}}_{t,i},\mathbf{y}_i)}\). This setup also
naturally reflects the incremental elimination of uncertainty: at the
beginning of the conversation/sequence, there is barely any information
to predict the outcome, so the output probability should be around the
population frequencies of classes in training samples. As conversation
progresses, the output probability density will gradually grow sharper
and concentrate towards the true label. This observation is essential to
the value definition. Given the goal of value assessment, the
predictions themselves are side-effects. However a well-tested high
performance predictive model is not only beneficial to systems which can
directly utilize its predictions (e.g.~DST could be one of the tasks),
but also more convincing for practitioners to rely on its value
estimations.

\hypertarget{value-function}{%
\subsection{Value Function}\label{value-function}}

In \protect\hyperlink{problem-formulation-and-data}{Section 2.1} we
discussed two kinds of tasks, one with a preferred outcome, the other
requiring confidence in prediction. For the three tasks, no-recontact
belongs to the former since we pursue a high probability for it, so the
value for this aspect can be simply set as
\(v^{(\text{no-recon})} = P(\text{no recontact})\). For issues and
actions, there is no preference on one class over the others. Even for
the binary actions, there is no preference on \emph{act} versus \emph{do
nothing} - many CS contacts end up to be purely informational or
educational. For them, we care about how much information the model has
gained during the conversation to yield confidence. We define this kind
of general information gain and confidence value as the KL-divergence
between the prediction and the non-informative distribution:
\[ V(p) \coloneqq D_{\text{KL}}(p\parallel p_0) = H(p,p_0) - H(p) \]
where \(p\) is the model's predicted distribution, \(H(\cdot,\cdot)\)
the cross-entropy function and \(H(\cdot)\) the entropy function.
\(p_0\) is the constant population empirical distribution of the target
across all training samples, as the best guess in case there is no
information provided at all at testing time. Intuitively, negative
entropy \(-H(p)\) measures the confidence or sharpness in the
prediction. The cross-entropy term measures the gap between the \(p\)
and \(p_0\), namely how far the prediction is from the non-informative
one. In summary, high \(V(p)\) means the model has obtained abundant
knowledge to make confident predictions.

Furthermore, for prediction \(p_t\) generated at the \(t\)th token in
the dialog sequences, \(V(p_t) - V(p_{t-k})\) captures the value change
caused by the \(k\) tokens in between. This yields a powerful approach
to evaluate the \emph{reward} or importance of any sentences in the
conversation\footnote{The value gain \(\Delta V\) is chosen instead of
  \(D_{\text{KL}}(p_t\parallel p_{t-k})\), for the sake of computation,
  storage and conceptual simplicity with the unified baseline \(p_0\).
  KL divergence does not satisfy triangle inequality so using it to
  compare any token pairs would cause inconsistency.}. In general,
\(R(t) \coloneqq V(p_{t+1}) - V(p_{t})\) enables token- or
sentence-level reward shaping for reinforcement learning agents, in the
case that there is no strong reward signals but some relevant while
indirect dialog-level attributes. The main idea is: even not directly
goal-related, more knowledge and less uncertainty is always a better
state for any solution-driven agents. Note for goal-oriented dialog, it
is also desirable to achieve a high-level of confidence \emph{early}, so
the turn-level value series \(\{V(p_{t})\}_t\) contain information to
measure the conversation progress efficiency, as detailed in
\protect\hyperlink{high-value-sentences-and-dialogs-for-contact-understanding}{Section
3.3}.

For human inspection, the multi-dimensional value vector
\([v^{(\text{issue})},v^{(\text{action})},v^{(\text{no-recon})}]\) is
helpful to analyze and interpret the bot or agent behaviors (e.g.~`bot
said this to increase no-recontact probability'). For automated
solutions, the value vector has to be collapsed into a scalar, to enable
ranking and optimization. We recommend weighting the different
dimensions based on the application with domain knowledge. For
demonstration, though the different dimensions in the value vector are
not on the same scale, we define the collapsed value as a simple
weighted average:
\[ v = \alpha v^{(\text{issue})} + \beta v^{(\text{action})} + v^{(\text{no-recon})} \]
where \(\alpha\) and \(\beta\) are normalizing constants to be set
empirically, so that the scales of value contribution are similar across
the tasks\footnote{Another candidate could be using the sum of empirical
  quantiles of all different values.This would be better in terms of a
  uniform normalization, while also being more risky in overlooking
  certain patterns.}. Finally, although the cost aspect is not covered,
the general regression variant of value definition is briefly discussed
in \protect\hyperlink{appendix-a-value-for-a-regression-task}{Appendix
A} for completeness.

\hypertarget{applications}{%
\section{Applications}\label{applications}}

\begin{longtable}[]{@{}rllrl@{}}
\caption{10 from the 60 samples comparing the original top-1 response
from RG with the re-ranked top-1 after adding VP signals. Contexts are
not shown. Human evaluation on the actual winner and the reason are
listed, with the CSA's previous choice marked as
\emph{italic}.}\tabularnewline
\toprule
\begin{minipage}[b]{0.02\columnwidth}\raggedleft
ID\strut
\end{minipage} & \begin{minipage}[b]{0.05\columnwidth}\raggedright
Winner\strut
\end{minipage} & \begin{minipage}[b]{0.16\columnwidth}\raggedright
Reason\strut
\end{minipage} & \begin{minipage}[b]{0.05\columnwidth}\raggedleft
Model\strut
\end{minipage} & \begin{minipage}[b]{0.58\columnwidth}\raggedright
Top 1 Recommendation\strut
\end{minipage}\tabularnewline
\midrule
\endfirsthead
\toprule
\begin{minipage}[b]{0.02\columnwidth}\raggedleft
ID\strut
\end{minipage} & \begin{minipage}[b]{0.05\columnwidth}\raggedright
Winner\strut
\end{minipage} & \begin{minipage}[b]{0.16\columnwidth}\raggedright
Reason\strut
\end{minipage} & \begin{minipage}[b]{0.05\columnwidth}\raggedleft
Model\strut
\end{minipage} & \begin{minipage}[b]{0.58\columnwidth}\raggedright
Top 1 Recommendation\strut
\end{minipage}\tabularnewline
\midrule
\endhead
\begin{minipage}[t]{0.02\columnwidth}\raggedleft
1\strut
\end{minipage} & \begin{minipage}[t]{0.05\columnwidth}\raggedright
VP\strut
\end{minipage} & \begin{minipage}[t]{0.16\columnwidth}\raggedright
informative\strut
\end{minipage} & \begin{minipage}[t]{0.05\columnwidth}\raggedleft
RG\strut
\end{minipage} & \begin{minipage}[t]{0.58\columnwidth}\raggedright
Sure.\strut
\end{minipage}\tabularnewline
\begin{minipage}[t]{0.02\columnwidth}\raggedleft
\strut
\end{minipage} & \begin{minipage}[t]{0.05\columnwidth}\raggedright
\strut
\end{minipage} & \begin{minipage}[t]{0.16\columnwidth}\raggedright
\strut
\end{minipage} & \begin{minipage}[t]{0.05\columnwidth}\raggedleft
VP\strut
\end{minipage} & \begin{minipage}[t]{0.58\columnwidth}\raggedright
\emph{Refund issued successfully.}\strut
\end{minipage}\tabularnewline
\begin{minipage}[t]{0.02\columnwidth}\raggedleft
2\strut
\end{minipage} & \begin{minipage}[t]{0.05\columnwidth}\raggedright
VP\strut
\end{minipage} & \begin{minipage}[t]{0.16\columnwidth}\raggedright
informative\strut
\end{minipage} & \begin{minipage}[t]{0.05\columnwidth}\raggedleft
RG\strut
\end{minipage} & \begin{minipage}[t]{0.58\columnwidth}\raggedright
Sorry to hear that the item was damaged.\strut
\end{minipage}\tabularnewline
\begin{minipage}[t]{0.02\columnwidth}\raggedleft
\strut
\end{minipage} & \begin{minipage}[t]{0.05\columnwidth}\raggedright
\strut
\end{minipage} & \begin{minipage}[t]{0.16\columnwidth}\raggedright
\strut
\end{minipage} & \begin{minipage}[t]{0.05\columnwidth}\raggedleft
VP\strut
\end{minipage} & \begin{minipage}[t]{0.58\columnwidth}\raggedright
\emph{In this case, I can process the refund for you.}\strut
\end{minipage}\tabularnewline
\begin{minipage}[t]{0.02\columnwidth}\raggedleft
3\strut
\end{minipage} & \begin{minipage}[t]{0.05\columnwidth}\raggedright
VP\strut
\end{minipage} & \begin{minipage}[t]{0.16\columnwidth}\raggedright
follows context\strut
\end{minipage} & \begin{minipage}[t]{0.05\columnwidth}\raggedleft
RG\strut
\end{minipage} & \begin{minipage}[t]{0.58\columnwidth}\raggedright
Are you able to print your own label?\strut
\end{minipage}\tabularnewline
\begin{minipage}[t]{0.02\columnwidth}\raggedleft
\strut
\end{minipage} & \begin{minipage}[t]{0.05\columnwidth}\raggedright
\strut
\end{minipage} & \begin{minipage}[t]{0.16\columnwidth}\raggedright
\strut
\end{minipage} & \begin{minipage}[t]{0.05\columnwidth}\raggedleft
VP\strut
\end{minipage} & \begin{minipage}[t]{0.58\columnwidth}\raggedright
\emph{I've successfully created a return label for you.}\strut
\end{minipage}\tabularnewline
\begin{minipage}[t]{0.02\columnwidth}\raggedleft
4\strut
\end{minipage} & \begin{minipage}[t]{0.05\columnwidth}\raggedright
VP\strut
\end{minipage} & \begin{minipage}[t]{0.16\columnwidth}\raggedright
follows context\strut
\end{minipage} & \begin{minipage}[t]{0.05\columnwidth}\raggedleft
RG\strut
\end{minipage} & \begin{minipage}[t]{0.58\columnwidth}\raggedright
Anything else needs to be taken care of?\strut
\end{minipage}\tabularnewline
\begin{minipage}[t]{0.02\columnwidth}\raggedleft
\strut
\end{minipage} & \begin{minipage}[t]{0.05\columnwidth}\raggedright
\strut
\end{minipage} & \begin{minipage}[t]{0.16\columnwidth}\raggedright
\strut
\end{minipage} & \begin{minipage}[t]{0.05\columnwidth}\raggedleft
VP\strut
\end{minipage} & \begin{minipage}[t]{0.58\columnwidth}\raggedright
\emph{Replacement successful.}\strut
\end{minipage}\tabularnewline
\begin{minipage}[t]{0.02\columnwidth}\raggedleft
5\strut
\end{minipage} & \begin{minipage}[t]{0.05\columnwidth}\raggedright
VP\strut
\end{minipage} & \begin{minipage}[t]{0.16\columnwidth}\raggedright
saved 2 turns\strut
\end{minipage} & \begin{minipage}[t]{0.05\columnwidth}\raggedleft
RG\strut
\end{minipage} & \begin{minipage}[t]{0.58\columnwidth}\raggedright
\emph{Great!}\strut
\end{minipage}\tabularnewline
\begin{minipage}[t]{0.02\columnwidth}\raggedleft
\strut
\end{minipage} & \begin{minipage}[t]{0.05\columnwidth}\raggedright
\strut
\end{minipage} & \begin{minipage}[t]{0.16\columnwidth}\raggedright
\strut
\end{minipage} & \begin{minipage}[t]{0.05\columnwidth}\raggedleft
VP\strut
\end{minipage} & \begin{minipage}[t]{0.58\columnwidth}\raggedright
\ldots you prefer refund in original payment method or gift card?\strut
\end{minipage}\tabularnewline
\begin{minipage}[t]{0.02\columnwidth}\raggedleft
6\strut
\end{minipage} & \begin{minipage}[t]{0.05\columnwidth}\raggedright
RG\strut
\end{minipage} & \begin{minipage}[t]{0.16\columnwidth}\raggedright
question repeats\strut
\end{minipage} & \begin{minipage}[t]{0.05\columnwidth}\raggedleft
RG\strut
\end{minipage} & \begin{minipage}[t]{0.58\columnwidth}\raggedright
\emph{I have issued a refund in original payment method
method\ldots{}}\strut
\end{minipage}\tabularnewline
\begin{minipage}[t]{0.02\columnwidth}\raggedleft
\strut
\end{minipage} & \begin{minipage}[t]{0.05\columnwidth}\raggedright
\strut
\end{minipage} & \begin{minipage}[t]{0.16\columnwidth}\raggedright
\strut
\end{minipage} & \begin{minipage}[t]{0.05\columnwidth}\raggedleft
VP\strut
\end{minipage} & \begin{minipage}[t]{0.58\columnwidth}\raggedright
Quick question: a refund in gift card or original payment\ldots{}\strut
\end{minipage}\tabularnewline
\begin{minipage}[t]{0.02\columnwidth}\raggedleft
7\strut
\end{minipage} & \begin{minipage}[t]{0.05\columnwidth}\raggedright
RG\strut
\end{minipage} & \begin{minipage}[t]{0.16\columnwidth}\raggedright
offer too early\strut
\end{minipage} & \begin{minipage}[t]{0.05\columnwidth}\raggedleft
RG\strut
\end{minipage} & \begin{minipage}[t]{0.58\columnwidth}\raggedright
\emph{Yes.}\strut
\end{minipage}\tabularnewline
\begin{minipage}[t]{0.02\columnwidth}\raggedleft
\strut
\end{minipage} & \begin{minipage}[t]{0.05\columnwidth}\raggedright
\strut
\end{minipage} & \begin{minipage}[t]{0.16\columnwidth}\raggedright
\strut
\end{minipage} & \begin{minipage}[t]{0.05\columnwidth}\raggedleft
VP\strut
\end{minipage} & \begin{minipage}[t]{0.58\columnwidth}\raggedright
Refund to original payment method or gift card?\strut
\end{minipage}\tabularnewline
\begin{minipage}[t]{0.02\columnwidth}\raggedleft
8\strut
\end{minipage} & \begin{minipage}[t]{0.05\columnwidth}\raggedright
both\strut
\end{minipage} & \begin{minipage}[t]{0.16\columnwidth}\raggedright
same meaning\strut
\end{minipage} & \begin{minipage}[t]{0.05\columnwidth}\raggedleft
RG\strut
\end{minipage} & \begin{minipage}[t]{0.58\columnwidth}\raggedright
Refund to original payment method or gift card?\strut
\end{minipage}\tabularnewline
\begin{minipage}[t]{0.02\columnwidth}\raggedleft
\strut
\end{minipage} & \begin{minipage}[t]{0.05\columnwidth}\raggedright
\strut
\end{minipage} & \begin{minipage}[t]{0.16\columnwidth}\raggedright
\strut
\end{minipage} & \begin{minipage}[t]{0.05\columnwidth}\raggedleft
VP\strut
\end{minipage} & \begin{minipage}[t]{0.58\columnwidth}\raggedright
\emph{Quick question: a refund in gift card or original
payment\ldots{}}\strut
\end{minipage}\tabularnewline
\begin{minipage}[t]{0.02\columnwidth}\raggedleft
9\strut
\end{minipage} & \begin{minipage}[t]{0.05\columnwidth}\raggedright
both\strut
\end{minipage} & \begin{minipage}[t]{0.16\columnwidth}\raggedright
clarification\strut
\end{minipage} & \begin{minipage}[t]{0.05\columnwidth}\raggedleft
RG\strut
\end{minipage} & \begin{minipage}[t]{0.58\columnwidth}\raggedright
May I know the date, amount and last 4 digits of the card\ldots{}\strut
\end{minipage}\tabularnewline
\begin{minipage}[t]{0.02\columnwidth}\raggedleft
\strut
\end{minipage} & \begin{minipage}[t]{0.05\columnwidth}\raggedright
\strut
\end{minipage} & \begin{minipage}[t]{0.16\columnwidth}\raggedright
\strut
\end{minipage} & \begin{minipage}[t]{0.05\columnwidth}\raggedleft
VP\strut
\end{minipage} & \begin{minipage}[t]{0.58\columnwidth}\raggedright
\emph{Let me check it for you.}\strut
\end{minipage}\tabularnewline
\begin{minipage}[t]{0.02\columnwidth}\raggedleft
10\strut
\end{minipage} & \begin{minipage}[t]{0.05\columnwidth}\raggedright
both\strut
\end{minipage} & \begin{minipage}[t]{0.16\columnwidth}\raggedright
non-informative\strut
\end{minipage} & \begin{minipage}[t]{0.05\columnwidth}\raggedleft
RG\strut
\end{minipage} & \begin{minipage}[t]{0.58\columnwidth}\raggedright
\emph{Thanks for understanding.}\strut
\end{minipage}\tabularnewline
\begin{minipage}[t]{0.02\columnwidth}\raggedleft
\strut
\end{minipage} & \begin{minipage}[t]{0.05\columnwidth}\raggedright
\strut
\end{minipage} & \begin{minipage}[t]{0.16\columnwidth}\raggedright
\strut
\end{minipage} & \begin{minipage}[t]{0.05\columnwidth}\raggedleft
VP\strut
\end{minipage} & \begin{minipage}[t]{0.58\columnwidth}\raggedright
You're welcome.\strut
\end{minipage}\tabularnewline
\bottomrule
\end{longtable}

We assess the performance of Value Profiler on an Amazon Customer
Service Contacts dataset for three general applications. 1.6MM text chat
contacts (sensitive fields removed) for US marketplace are sampled with
the corresponding targets described in
\protect\hyperlink{problem-formulation-and-data}{Section 2.1}, and split
90/10 into training and testing set. For quantitative evaluation, we
investigate how VP can benefit ruled-based dialog trees as well as
state-of-the-art dialog generators (WF and RG; discussed in
\protect\hyperlink{background-and-motivation}{Section 1}) by comparing
metrics with or without VP's input in carefully designed offline, and
small scale online studies\footnote{Larger scale online tests are
  scheduled.}. For qualitative evaluation, we show that the value
estimates align with human judgments, by sampling non-cherry-picking
random examples as test sets.

\textbf{Model Details} The scrubbed transcript of each contact is
tokenized by the GPT2 tokenizer (Byte-Pair Encoding) into a
variable-length list of tokens. For training, a mini-batch of 32 samples
is generated, each being a contact with 512 tokens. Contact shorter than
512 tokens are padded on the right, and those longer are randomly
(different for each epoch) sub-sequenced into 512 tokens. The mini-batch
goes into the causal encoder, the encoder outputs token-level
embeddings, followed by a weight-shared hidden layer of size 128 with
ReLU activation. Finally the model outputs the softmax or sigmoid
probabilities for each class in each task. The targets are replicated
across all tokens, and linked to the outputs with a cross-entropy loss.
For the pre-trained GPT2 encoder, we use the Tensorflow implementation
from \href{https://arxiv.org/pdf/1910.03771.pdf}{Wolf et al., 2019} and
fine-tune for 3 epochs; for CNN/RNN encoder, the model is trained from
scratch with 100 epochs. Adam optimizer with learning rate 0.0001 is
used for fine-tuning and 0.001 for full training. The output values are
collapsed as described in the previous section then normalized. To
stabilize variance, sentence-level value is computed as the moving
average of 7 token-level values centered at the last token of the
sentence if evaluating offline, or 4 preceding token-level values if
predicting online. The model architecture and training setting is
intended to be a bare minimal implementation for the functionality
without hyperparameter tuning.

\hypertarget{prediction-accuracy}{%
\subsection{Prediction Accuracy}\label{prediction-accuracy}}

In this application we show VP has high accuracy in the predictive
tasks, as a piece of evidence to support our trust in its value
estimation. VP predicts the contact issue, the resolving actions and
recontact status, all of them can influence dialog management decisions.
There is no existing benchmarks for recontact and action prediction for
this dataset, but the WF system has an NLU component and outputs intent
predictions. We design the following experiment to compare the VP issue
(intent) prediction against \emph{the whole WF dialog system}, including
the NLU component and other deterministic routing rules. We sample 68111
contacts with issue codes filled by CSAs after the contact, and examine
the issue prediction by VP at the last WF sentence, right before CSA
joins, against the issue code as the ground truth. The benchmark is set
to be \emph{the actual last WF visited} in this contact. The idea is
that, the last WF (with its intent label) indicates the best guess from
the whole WF system back then about the customer issue, and the VP
prediction at the same time is a fair competitor, both compared to the
final label by CSA.

Accuracy comparison is shown in
\protect\hyperlink{prediction-accuracy-for-dialog-management}{Table 1},
grouped by some major intents and VP is better in most cases. The
accuracy advantage mainly comes from the most ambiguous intents
(\emph{live-help-request}, \emph{order-related}) and those with
confusing topics (\emph{unknown charge}, \emph{account}, \emph{prime}).
Though without previous benchmarks on this dataset, the metrics of
action and recontact predictions are considered as reasonable: depending
on predicting at which point during the conversation, the accuracy for
no-recontact prediction is from 72\% (non-informative) to 83\% (after
reading all 512 tokens); across action types, the action prediction
precision at the end of the contact is from 58\% to 94\%, and recall
from 42\% to 92\%. For reference, many actions have 99\%+ negative
labels. \textbf{Error analysis} was conducted with a random sample of
size 20 from each task where the prediction and label disagrees. By
human evaluation, VP predictions for 78.3\% (47 out of 60) of these
`wrong' cases are in fact more or equally appropriate compared to the
label. This means all above metrics are underestimates, and implies
potential applications in label imputation and correction.
\textbf{Calibration analysis} (not shown) indicates that for most tasks
except few issue categories, the output distribution is well calibrated.
\textbf{Volatility analysis} based on five replicated experiments with
random seeds show the accuracy fluctuations are within 1\%. The
\textbf{CNN/LSTM benchmarks} are both 10\%+ worse than GPT2 for all
tasks thus not further investigated. See
\protect\hyperlink{appendix-c-alternative-encoders-and-volatility-results}{Appendix
C} for detailed volatility and CNN/LSTM results.

\hypertarget{reward-signals-for-text-generation}{%
\subsection{Reward Signals for Text
Generation}\label{reward-signals-for-text-generation}}

There are various ways that VP can inject the reward signals into any
dialog generator: (1) post-process the recommendation list (i.e.~beam
search or retrieval results) and re-rank by the estimated rewards of
each suggestion; (2) pre-process to assign sample weights to the text
generator's training set by the estimated reward of target sentences;
(3) Off-line Reinforcement Learning: VP serves as a static critic to
provide rewards to the text generator training through policy gradient.
In this text we demonstrate the first two approaches for their
simplicity in integration with an existing dialog generator, and leave
(3) to future work.

\textbf{Re-Ranking} We sampled a separate set of 47078 contacts, from
some previous online experiments where CSAs were asked to read through
the suggestions from the text generator (RG), until spotting the first
applicable response then select it. Top 4 suggestions from RG are
gathered, each with a score. This score is either a negative
log-likelihood or a retrieval matching score, depending on the two types
of RG model (see \href{https://www.aclweb.org/anthology/N19-2007.pdf}{Lu
et al, 2018} for details). We gather scores from RG and reward estimates
from VP for each list, normalize both into \([0,1]\), then take the
average as the ensemble score representing both the likelihood and the
value of a response. The agent's turn acceptance rate (TAR-1; percentage
of top-1 recommendations from RG gets accepted/chosen by CSAs) is
considered as the gold standard to assess RG performance. Note it is not
fair to use the TAR-1 computed from the VP-influenced ranking and CSA's
choice \emph{back then}, because this is an offline dataset and the
perception bias caused by the original ranking would dominate. Instead,
we draw a random sample of size 30 from each of the two sets:
\textbf{(A)} CSA accepted RG's top-1 but VP has a different top-1;
\textbf{(B)} CSA did not accept RG's top-1 but VP re-ranks their choice
as top-1. Through this small scale human evaluation, the behavior
difference between the two models is analyzed below, with some examples
from this analysis listed in
\protect\hyperlink{reward-signals-for-text-generation}{Table 2}.
Results: \textbf{19} out of \textbf{30} (63\%) in set A and \textbf{11}
out of \textbf{30} (36\%) in set B are contacts that both top-1
suggestion are acceptable or have the same meaning; note the percentage
difference indicates the bias towards the original ranking. In the
remaining \textbf{11} contacts in set A, RG response is indeed better
than VP in \textbf{7} of them, all due to VP responses hurrying to a
solution that is either too early or out of context; for the other
\textbf{4} contacts, VP responses are more appropriate. In the remaining
\textbf{19} contacts in set B, VP response is indeed better than RG for
\emph{all of them}, because RG's original response was either
non-informative or out of contexts. This agrees with our expectation.

\textbf{Training Weights} A separate small-scale Japanese marketplace
online A/B experiment is conducted to compare RG (control) and the same
RG trained with VP induced sample weights (treatment). The weights are
estimated target rewards, normlized to \([0,1]\). 4 suggestions are
generated by each model for a test set of 500 contacts and sent to
Japanese annotators to choose from. Results: treatment improves (\%
relative to control; with bootstrap 90\% Confidence Interval) MRR (mean
reciprocal rank) by \textbf{4.04\% (-1.6\% \(\sim\) 9.61\%)} and TAR-1
by \textbf{8.64\% (-3.29\% \(\sim\) 20.39\%)}. Statistical
insignificance should come from limited sample size and noisy
annotation, as the CIs are pretty wide. A larger size study is planned.
Error analysis showed similar qualitative result as the re-ranking.

\hypertarget{high-value-sentences-and-dialogs-for-contact-understanding}{%
\subsection{High Value Sentences and Dialogs for Contact
Understanding}\label{high-value-sentences-and-dialogs-for-contact-understanding}}

\begin{longtable}[]{@{}llrl@{}}
\caption{Sampled sentences with top positive, negative and near-zero
rewards (\(\Delta v\)) for different aspects. The reward numbers are
scaled and the customer turns have been rephrased.}\tabularnewline
\toprule
\begin{minipage}[b]{0.07\columnwidth}\raggedright
Value\strut
\end{minipage} & \begin{minipage}[b]{0.07\columnwidth}\raggedright
Tier\strut
\end{minipage} & \begin{minipage}[b]{0.08\columnwidth}\raggedleft
\(\Delta v\)\strut
\end{minipage} & \begin{minipage}[b]{0.66\columnwidth}\raggedright
Sentence (A: agent, B: bot, C: customer)\strut
\end{minipage}\tabularnewline
\midrule
\endfirsthead
\toprule
\begin{minipage}[b]{0.07\columnwidth}\raggedright
Value\strut
\end{minipage} & \begin{minipage}[b]{0.07\columnwidth}\raggedright
Tier\strut
\end{minipage} & \begin{minipage}[b]{0.08\columnwidth}\raggedleft
\(\Delta v\)\strut
\end{minipage} & \begin{minipage}[b]{0.66\columnwidth}\raggedright
Sentence (A: agent, B: bot, C: customer)\strut
\end{minipage}\tabularnewline
\midrule
\endhead
\begin{minipage}[t]{0.07\columnwidth}\raggedright
action\strut
\end{minipage} & \begin{minipage}[t]{0.07\columnwidth}\raggedright
positive\strut
\end{minipage} & \begin{minipage}[t]{0.08\columnwidth}\raggedleft
0.263\strut
\end{minipage} & \begin{minipage}[t]{0.66\columnwidth}\raggedright
C: I forgot my password, so I cannot change it\strut
\end{minipage}\tabularnewline
\begin{minipage}[t]{0.07\columnwidth}\raggedright
action\strut
\end{minipage} & \begin{minipage}[t]{0.07\columnwidth}\raggedright
positive\strut
\end{minipage} & \begin{minipage}[t]{0.08\columnwidth}\raggedleft
0.248\strut
\end{minipage} & \begin{minipage}[t]{0.66\columnwidth}\raggedright
A: I will go ahead and cancel the order for stuck shipment. Okay?\strut
\end{minipage}\tabularnewline
\begin{minipage}[t]{0.07\columnwidth}\raggedright
action\strut
\end{minipage} & \begin{minipage}[t]{0.07\columnwidth}\raggedright
negative\strut
\end{minipage} & \begin{minipage}[t]{0.08\columnwidth}\raggedleft
-0.112\strut
\end{minipage} & \begin{minipage}[t]{0.66\columnwidth}\raggedright
A: I see you have been refunded for this item\strut
\end{minipage}\tabularnewline
\begin{minipage}[t]{0.07\columnwidth}\raggedright
issue\strut
\end{minipage} & \begin{minipage}[t]{0.07\columnwidth}\raggedright
positive\strut
\end{minipage} & \begin{minipage}[t]{0.08\columnwidth}\raggedleft
0.684\strut
\end{minipage} & \begin{minipage}[t]{0.66\columnwidth}\raggedright
C: I want to cancel my amazon music subscription, immediately\strut
\end{minipage}\tabularnewline
\begin{minipage}[t]{0.07\columnwidth}\raggedright
issue\strut
\end{minipage} & \begin{minipage}[t]{0.07\columnwidth}\raggedright
positive\strut
\end{minipage} & \begin{minipage}[t]{0.08\columnwidth}\raggedleft
0.537\strut
\end{minipage} & \begin{minipage}[t]{0.66\columnwidth}\raggedright
C: \ldots I can't post reviews about some products that I've
purchased\strut
\end{minipage}\tabularnewline
\begin{minipage}[t]{0.07\columnwidth}\raggedright
issue\strut
\end{minipage} & \begin{minipage}[t]{0.07\columnwidth}\raggedright
negative\strut
\end{minipage} & \begin{minipage}[t]{0.08\columnwidth}\raggedleft
-0.213\strut
\end{minipage} & \begin{minipage}[t]{0.66\columnwidth}\raggedright
C: No, that's all\strut
\end{minipage}\tabularnewline
\begin{minipage}[t]{0.07\columnwidth}\raggedright
recontact\strut
\end{minipage} & \begin{minipage}[t]{0.07\columnwidth}\raggedright
positive\strut
\end{minipage} & \begin{minipage}[t]{0.08\columnwidth}\raggedleft
0.625\strut
\end{minipage} & \begin{minipage}[t]{0.66\columnwidth}\raggedright
C: It is too late for a phone call. I can do another chat
tomorrow\ldots{}\strut
\end{minipage}\tabularnewline
\begin{minipage}[t]{0.07\columnwidth}\raggedright
recontact\strut
\end{minipage} & \begin{minipage}[t]{0.07\columnwidth}\raggedright
positive\strut
\end{minipage} & \begin{minipage}[t]{0.08\columnwidth}\raggedleft
0.416\strut
\end{minipage} & \begin{minipage}[t]{0.66\columnwidth}\raggedright
C: I prefer online chat, since english is my second language.\strut
\end{minipage}\tabularnewline
\begin{minipage}[t]{0.07\columnwidth}\raggedright
recontact\strut
\end{minipage} & \begin{minipage}[t]{0.07\columnwidth}\raggedright
negative\strut
\end{minipage} & \begin{minipage}[t]{0.08\columnwidth}\raggedleft
-0.472\strut
\end{minipage} & \begin{minipage}[t]{0.66\columnwidth}\raggedright
A: Let me connect you to the carrier, may I have your phone
number?\strut
\end{minipage}\tabularnewline
\begin{minipage}[t]{0.07\columnwidth}\raggedright
total\strut
\end{minipage} & \begin{minipage}[t]{0.07\columnwidth}\raggedright
positive\strut
\end{minipage} & \begin{minipage}[t]{0.08\columnwidth}\raggedleft
0.542\strut
\end{minipage} & \begin{minipage}[t]{0.66\columnwidth}\raggedright
A: I'll escalate to the carrier..rescheduling the delivery at the
earliest..\strut
\end{minipage}\tabularnewline
\begin{minipage}[t]{0.07\columnwidth}\raggedright
total\strut
\end{minipage} & \begin{minipage}[t]{0.07\columnwidth}\raggedright
positive\strut
\end{minipage} & \begin{minipage}[t]{0.08\columnwidth}\raggedleft
0.508\strut
\end{minipage} & \begin{minipage}[t]{0.66\columnwidth}\raggedright
B: Looks like this item should have been delivered by Thu, Feb 14.\strut
\end{minipage}\tabularnewline
\begin{minipage}[t]{0.07\columnwidth}\raggedright
total\strut
\end{minipage} & \begin{minipage}[t]{0.07\columnwidth}\raggedright
zero\strut
\end{minipage} & \begin{minipage}[t]{0.08\columnwidth}\raggedleft
0.001\strut
\end{minipage} & \begin{minipage}[t]{0.66\columnwidth}\raggedright
A: Yes, I can check the details.\strut
\end{minipage}\tabularnewline
\begin{minipage}[t]{0.07\columnwidth}\raggedright
total\strut
\end{minipage} & \begin{minipage}[t]{0.07\columnwidth}\raggedright
zero\strut
\end{minipage} & \begin{minipage}[t]{0.08\columnwidth}\raggedleft
0\strut
\end{minipage} & \begin{minipage}[t]{0.66\columnwidth}\raggedright
B: Hi! It's Amazon's messaging assistant again.\strut
\end{minipage}\tabularnewline
\begin{minipage}[t]{0.07\columnwidth}\raggedright
total\strut
\end{minipage} & \begin{minipage}[t]{0.07\columnwidth}\raggedright
zero\strut
\end{minipage} & \begin{minipage}[t]{0.08\columnwidth}\raggedleft
-0.001\strut
\end{minipage} & \begin{minipage}[t]{0.66\columnwidth}\raggedright
C: An item I ordered\strut
\end{minipage}\tabularnewline
\begin{minipage}[t]{0.07\columnwidth}\raggedright
total\strut
\end{minipage} & \begin{minipage}[t]{0.07\columnwidth}\raggedright
negative\strut
\end{minipage} & \begin{minipage}[t]{0.08\columnwidth}\raggedleft
-0.485\strut
\end{minipage} & \begin{minipage}[t]{0.66\columnwidth}\raggedright
A: To resolve the issue, I need to connect you to our specialist
team\strut
\end{minipage}\tabularnewline
\bottomrule
\end{longtable}

\begin{figure}
\centering
\includegraphics[width=2.91667in,height=1.51042in]{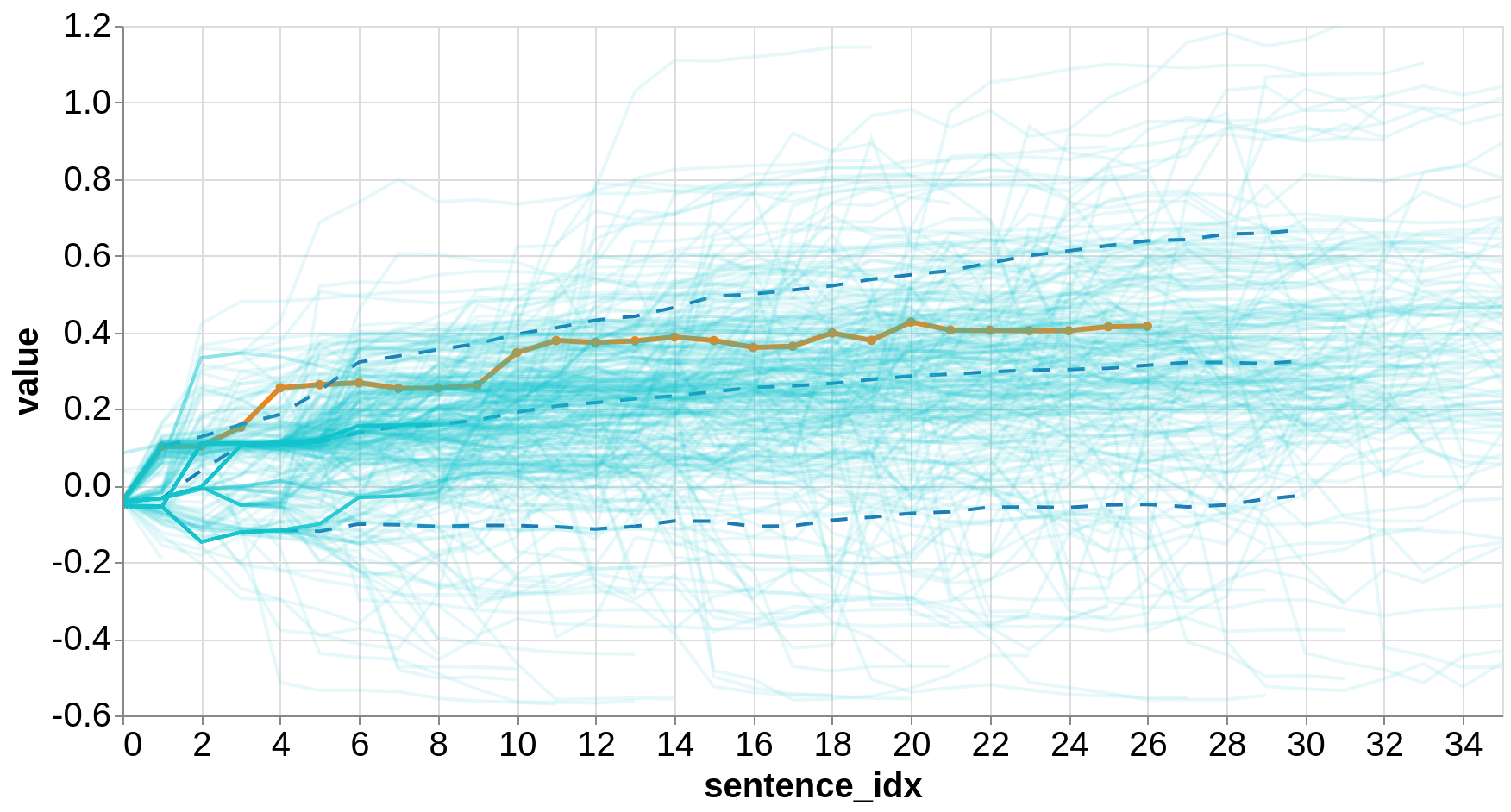}
\caption{Turn-level Value progress curve distribution. Each light blue
solid line shows 1 of 300 random chats; the three blue dashed lines are
the P10/P50/P90 curves computed based on 30000 random chats. A random
high-value (above P50) sample is highlighted in orange. See
\protect\hyperlink{appendix-b-an-example-high-value-contact}{Appendix B}
for the transcript of the sample.}
\end{figure}

Apart from automation-related applications, Value Profiler enables
scalable CS contacts understanding and evaluation, to minimize the
effort in contact reading by specialists or sending contact survey with
sparse response. VP automatically quantifies the conversation progress,
and labels different aspects of the contact up to word-level
granularity. For any part of the conversation, VP can quantify its
contribution to the multi-aspect success: reducing recontacts, acquiring
information, and reducing costs. This contributions could be positive,
negative or insignificant. In
\protect\hyperlink{high-value-sentences-and-dialogs-for-contact-understanding}{Table
3}, some randomly sampled sentences are listed, representing top
positive, negative and near-zero reward for each aspect. Extracting
high-value sentences not only provides insights and highlights of the
contact, but also can assist the design of dialog system actions: if
uncertainties are observed in some aspect, the system can select a
question that historically results in huge information gain in the
subsequent turn regarding to that aspect. We leave this as future work.

To evaluate whole dialog quality, we need more than turn-level rewards.
As shown in the upper part of
\protect\hyperlink{value-profiler-predictive-model}{Figure 2}, the ideal
value progress starts from zero (non-informative), quickly climbs up
(acquire/provide information) and saturates at a high level (resolved
and satisfied). To quantify this intuition, we propose to evaluate
dialog quality by comparing its \emph{value progress curve} (value vs
\#turns) with the \emph{quantile curves} of reference coordinated. For
all the value curves from a subset of contacts, the quantile curves are
defined as the connected point-wise empirical quantiles for each turn.
For example, the P90 quantile curve consists of \(\hat{v}^{(0.9)}_{t}\)
for each \(t\), so that \(P(v_{i,t} \leq \hat{v}^{(0.9)}_{t}) = 0.9\)
across the all samples. This can be interpreted as the P90 curve is
higher than 90\% of the contact curves. This quantile comparison yields
a relative criteria to define high/low-value contacts and quantify its
progress. In
\protect\hyperlink{high-value-sentences-and-dialogs-for-contact-understanding}{Figure
3}, the value curves of 300 random contacts are plotted against the
P10/P50/P90 quantile curves calculated from 30000 random contacts. The
orange curve is a random sample drawn between the P50 and P90, so its
quality is above median. The transcript of this example (with customer
turns masked) can be found in
\protect\hyperlink{appendix-b-an-example-high-value-contact}{Appendix
B}: the CSA approached the issue politely, proactively and timely. Other
examples below the P50 curve (not shown) contains long-winded
conversations, some of them showing customers trapped in WF loops. The
quantile curve not only provides an automatic metric to evaluate
historical contact quality and compare dialog models or human agents
performance, it can also make online decisions when a conversation with
a bot is trapped in the low-value area for too long. VP would consider
it as a \emph{bot failure}, and promptly suggest a CSA to take over and
ensure customer experience.

\hypertarget{related-work}{%
\section{Related Work}\label{related-work}}

\textbf{Information Gain used in RL and Dialog} Information gain as a
signal in Reinforcement Learning is not new, but mostly used as a
`curiosity' measure for exploration-exploitation in the next-state
prediction tasks (\href{https://arxiv.org/pdf/1810.12894.pdf}{Burda et
al., 2018}; \href{https://arxiv.org/pdf/1810.12162.pdf}{Shyam et al.,
2019}). \href{https://www.aclweb.org/anthology/P19-1101/}{Peyrard, 2019}
introduced some information-theoretic framework to evaluate the
importance in text summarization. In the task-oriented dialog domain,
\href{https://arxiv.org/abs/1911.03598}{Yu et al, 2019} and
\href{https://arxiv.org/abs/1907.12021}{Shukla et al, 2019} uses
information gain to select which question to ask. Our work differs in
combining multi-aspect reward signals in a much more ambiguous CS dialog
setting, and provides a richer set of evaluation methods for both bot
and human conversation progresses.

\textbf{Reward Design for Text Generation} Using RL tricks to complement
Maximum Likelihood learning for sequence generation has been well
investigated. \href{https://arxiv.org/pdf/1511.06732.pdf}{Ranzato et
al., 2015} and \href{https://arxiv.org/pdf/1607.07086.pdf}{Bahdanau et
al., 2016} applied respectively REINFORCE and actor-critic approach to
incorporate whole-sequence rewards into seq2seq training. The reward
signal they used is BLEU and ROUGE, which are simple text statistics to
compute on any text generation tasks but not far from likelihood.
\href{https://arxiv.org/pdf/1609.05473.pdf}{Yu et al., 2016} and
\href{http://papers.nips.cc/paper/6908-adversarial-ranking-for-language-generation.pdf}{Lin
et al., 2017} used adversarial training with discriminator or ranker
loss as reward signals for policy gradients.
\href{https://arxiv.org/pdf/1606.01541.pdf}{Li et al., 2016} designed a
couple of heuristic rewards specifically for dialog generation, all
based on certain transformations of context log probabilities within the
texts. On the other end of spectrum, reward signals \emph{external} to
texts can be obtained much more expensively by either dedicated human
labeling (\href{https://arxiv.org/pdf/1702.03334.pdf}{Kandasamy et al.,
2017}; \href{https://arxiv.org/pdf/1804.06512.pdf}{Liu et al., 2018}),
or by restricting to specific dialog applications with well-defined
strong signals such as successful API calls for online reservation
(\href{https://arxiv.org/pdf/1606.02560.pdf}{Zhao and Eskenazi, 2016};
\href{https://arxiv.org/abs/1609.00777}{Dhingra et al., 2016}).
\href{https://arxiv.org/abs/1712.02838}{Zhou et al., 2017} is the
closest to our work and their reward design is based on both BLEU scores
and errors in predicting turn-level action API call slots. However,
turn-aligned signals are still required. Also the prediction error is
not available at test time, so real-time evaluation and monitoring is
impossible. The value profiler reward design differs from above, by
using external dialog-level weak signals.

\hypertarget{conclusion-and-future-work}{%
\section{Conclusion and Future Work}\label{conclusion-and-future-work}}

We presented Value Profiler, a deep learning model that profiles the
multi-dimensional success of goal-oriented conversations, trained by
only weak dialog-level signals. The dialog evaluation from VP can help
contact understanding and bot/agent performance assessment. For
injecting reward signals into text generation, we showed that the simple
and practical pre- or post-processing methods result in improvements,
however training with policy gradient methods is a formal choice to be
investigated. The dialog generation task can also be added to the task
list for joint optimization, leading to an end-to-end value-oriented
conversational agent.

\hypertarget{reference}{%
\section*{Reference}\label{reference}}
\addcontentsline{toc}{section}{Reference}

Bahdanau, Dzmitry, Philemon Brakel, Kelvin Xu, Anirudh Goyal, Ryan Lowe,
Joelle Pineau, Aaron Courville, and Yoshua Bengio. ``An actor-critic
algorithm for sequence prediction.'' arXiv preprint arXiv:1607.07086
(2016).

Burda, Yuri, Harrison Edwards, Amos Storkey, and Oleg Klimov.
``Exploration by random network distillation.'' arXiv preprint
arXiv:1810.12894 (2018).

Henderson, Matthew. ``Machine learning for dialog state tracking: A
review.'' (2015).

Holtzman, Ari, Jan Buys, Maxwell Forbes, and Yejin Choi. ``The curious
case of neural text degeneration.'' arXiv preprint arXiv:1904.09751
(2019).

Kandasamy, Kirthevasan, Yoram Bachrach, Ryota Tomioka, Daniel Tarlow,
and David Carter. ``Batch policy gradient methods for improving neural
conversation models.'' arXiv preprint arXiv:1702.03334 (2017).

Li, Jiwei, Michel Galley, Chris Brockett, Jianfeng Gao, and Bill Dolan.
``A diversity-promoting objective function for neural conversation
models.'' arXiv preprint arXiv:1510.03055 (2015).

Li, Jiwei, Will Monroe, Alan Ritter, Michel Galley, Jianfeng Gao, and
Dan Jurafsky. ``Deep reinforcement learning for dialogue generation.''
arXiv preprint arXiv:1606.01541 (2016).

Lin, Kevin, Dianqi Li, Xiaodong He, Zhengyou Zhang, and Ming-Ting Sun.
``Adversarial ranking for language generation.'' In Advances in Neural
Information Processing Systems, pp.~3155-3165. 2017.

Lipton, Zachary, Xiujun Li, Jianfeng Gao, Lihong Li, Faisal Ahmed, and
Li Deng. ``Bbq-networks: Efficient exploration in deep reinforcement
learning for task-oriented dialogue systems.'' In Thirty-Second AAAI
Conference on Artificial Intelligence. 2018.

Liu, Bing, Gokhan Tur, Dilek Hakkani-Tur, Pararth Shah, and Larry Heck.
``Dialogue learning with human teaching and feedback in end-to-end
trainable task-oriented dialogue systems.'' arXiv preprint
arXiv:1804.06512 (2018).

Lu, Yichao, Manisha Srivastava, Jared Kramer, Heba Elfardy, Andrea Kahn,
Song Wang, and Vikas Bhardwaj. ``Goal-Oriented End-to-End Conversational
Models with Profile Features in a Real-World Setting.'' In Proceedings
of the 2019 Conference of the North American Chapter of the Association
for Computational Linguistics: Human Language Technologies, Volume 2
(Industry Papers), pp.~48-55. 2019.

Oord, Aaron van den, Sander Dieleman, Heiga Zen, Karen Simonyan, Oriol
Vinyals, Alex Graves, Nal Kalchbrenner, Andrew Senior, and Koray
Kavukcuoglu. ``Wavenet: A generative model for raw audio.'' arXiv
preprint arXiv:1609.03499 (2016).

Peyrard, Maxime. ``A simple theoretical model of importance for
summarization.'' arXiv preprint arXiv:1801.08991 (2018).

Radford, Alec, Jeffrey Wu, Rewon Child, David Luan, Dario Amodei, and
Ilya Sutskever. ``Language models are unsupervised multitask learners.''
OpenAI Blog 1, no. 8 (2019): 9.

Ranzato, Marc'Aurelio, Sumit Chopra, Michael Auli, and Wojciech Zaremba.
``Sequence level training with recurrent neural networks.'' arXiv
preprint arXiv:1511.06732 (2015).

Sharma, Shikhar, Layla El Asri, Hannes Schulz, and Jeremie Zumer.
``Relevance of unsupervised metrics in task-oriented dialogue for
evaluating natural language generation.'' arXiv preprint
arXiv:1706.09799 (2017).

Shukla, Pushkar, Carlos Elmadjian, Richika Sharan, Vivek Kulkarni,
Matthew Turk, and William Yang Wang. ``What Should I Ask? Using
Conversationally Informative Rewards for Goal-Oriented Visual Dialog.''
arXiv preprint arXiv:1907.12021 (2019).

Shyam, Pranav, Wojciech Jaśkowski, and Faustino Gomez. ``Model-based
active exploration.'' arXiv preprint arXiv:1810.12162 (2018).

Welleck, Sean, Ilia Kulikov, Stephen Roller, Emily Dinan, Kyunghyun Cho,
and Jason Weston. ``Neural text generation with unlikelihood training.''
arXiv preprint arXiv:1908.04319 (2019).

Wolf, Thomas, Lysandre Debut, Victor Sanh, Julien Chaumond, Clement
Delangue, Anthony Moi, Pierric Cistac et al.. ``Transformers:
State-of-the-art Natural Language Processing.'' arXiv preprint
arXiv:1910.03771 (2019).

Yu, Lantao, Weinan Zhang, Jun Wang, and Yong Yu. ``Seqgan: Sequence
generative adversarial nets with policy gradient.'' In Thirty-First AAAI
Conference on Artificial Intelligence. 2017.

Yu, Lili, Howard Chen, Sida Wang, Yoav Artzi, and Tao Lei. ``Interactive
Classification by Asking Informative Questions.'' arXiv preprint
arXiv:1911.03598 (2019).

Zhang, Tianyi, Varsha Kishore, Felix Wu, Kilian Q. Weinberger, and Yoav
Artzi. ``Bertscore: Evaluating text generation with bert.'' arXiv
preprint arXiv:1904.09675 (2019).

Zhao, Wei, Maxime Peyrard, Fei Liu, Yang Gao, Christian M. Meyer, and
Steffen Eger. ``Moverscore: Text generation evaluating with
contextualized embeddings and earth mover distance.'' arXiv preprint
arXiv:1909.02622 (2019).

Zhou, Li, Kevin Small, Oleg Rokhlenko, and Charles Elkan. ``End-to-end
offline goal-oriented dialog policy learning via policy gradient.''
arXiv preprint arXiv:1712.02838 (2017).

\hypertarget{appendix-a-value-for-a-regression-task}{%
\section*{Appendix A Value for a Regression
Task}\label{appendix-a-value-for-a-regression-task}}
\addcontentsline{toc}{section}{Appendix A Value for a Regression Task}

For customer service contact evaluation, the cost aspect needs to be
modeled as a regression task. Lower costs is desirable so negative
predicted cost could be naively defined as the value directly. But there
is also the uncertainty side of the story, since a poor estimate of cost
with high volatility would invalidate the cost estimation number.
Entropy for continuous distributions is comparatively difficult to
compute, and depends on the choice of parametric distributions in a
neural net. Therefore, we recommend adding a nonparametric quantile
regression task. Let the model outputs two quantile estimates (e.g.~P10
\(\hat{y}^{(0.1)}\) and P90 \(\hat{y}^{(0.9)}\)) of the numeric target
(e.g.~cost \(y\) in dollars). Assume the model is calibrated (i.e.~P90
prediction does cover the true costs 90\% of the time:
\(P(y \leq \hat{y}^{(0.9)}) = 0.9\)), then the length of the prediction
interval \([\hat{y}^{(0.1)},\hat{y}^{(0.9)}]\) can be used to assess the
confidence or sharpness of the model prediction. Prediction interval
length is a general confidence measure for any numeric attributes. For
the particular cost case (lower the better), a negative quantile
estimate itself (e.g.~P90) could be directly used as the value,
considering both low-cost (effect-size) and high-confidence
(significance).

\begin{longtable}[]{@{}lrrrrrr@{}}
\caption{A lite comparison of different encoders and random seeds. All
losses are cross entropy on test set. Accuracy for recontact and issue
is computed at the 256th token (half way), and scaled by dividing that
of the CNN model. Action metric difference is small due to the highly
skewed distribution. The three GPT2 models shown are only fine-tuned for
one epoch, while the final model has 3 epochs and is more stable (not
shown).}\tabularnewline
\toprule
Encoder & total loss & recon loss & issue loss & action loss & recon
acc. & issue acc.\tabularnewline
\midrule
\endfirsthead
\toprule
Encoder & total loss & recon loss & issue loss & action loss & recon
acc. & issue acc.\tabularnewline
\midrule
\endhead
CNN & 5.235 & 0.791 & 2.522 & 0.0461 & 1.000 & 1.000\tabularnewline
LSTM & 5.381 & 0.788 & 2.586 & 0.0445 & 0.994 & 0.992\tabularnewline
GPT2 seed1 & 4.411 & 0.615 & 2.315 & 0.0421 & 1.178 &
1.152\tabularnewline
GPT2 seed2 & 4.448 & 0.612 & 2.343 & 0.0437 & 1.191 &
1.148\tabularnewline
GPT2 seed3 & 4.382 & 0.607 & 2.311 & 0.0428 & 1.203 &
1.168\tabularnewline
\bottomrule
\end{longtable}

\hypertarget{appendix-b-an-example-high-value-contact}{%
\section*{Appendix B An Example High Value
Contact}\label{appendix-b-an-example-high-value-contact}}
\addcontentsline{toc}{section}{Appendix B An Example High Value Contact}

See Figure 4 for the transcript of the highlighted contact in Figure 3.

\begin{figure}
\centering
\includegraphics[width=4.16667in,height=5in]{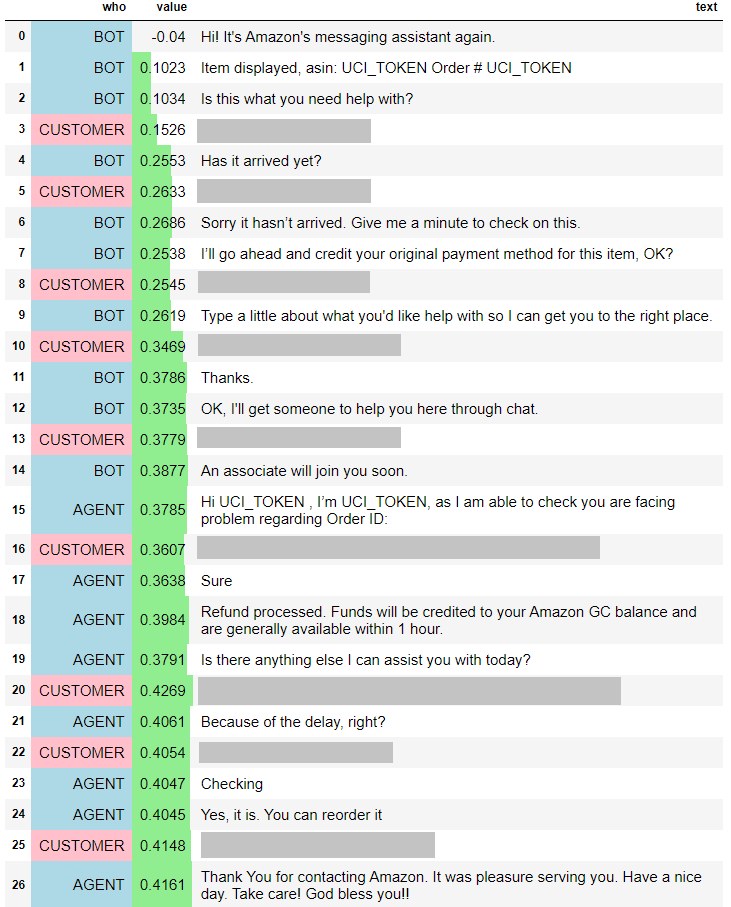}
\caption{The transcript of the orange example contact value curve in
Figure 3. Customer utterances are masked. The value jumps at highly
informative turns where the customer provides key information and the
CSA maintained the high value by solving the issue politely, proactively
and timely.}
\end{figure}

\hypertarget{appendix-c-alternative-encoders-and-volatility-results}{%
\section*{Appendix C Alternative Encoders and Volatility
Results}\label{appendix-c-alternative-encoders-and-volatility-results}}
\addcontentsline{toc}{section}{Appendix C Alternative Encoders and
Volatility Results}

See Table 4 for result metrics of LSTM/CNN encoders and volatility
analysis.

\end{document}